\documentclass[12pt]{article}

\usepackage[utf8]{inputenc}
\usepackage{amsmath}
\usepackage{amsfonts}
\usepackage{mathrsfs}
\usepackage{amssymb}
\usepackage{amsthm}
\usepackage{bm}
\usepackage{bbm}

\usepackage{libertine}
\usepackage{wrapfig}
\usepackage{graphicx}
\usepackage{multirow}
\usepackage{epstopdf}
\usepackage{multicol}
\usepackage{subfigure}

\usepackage{algorithm}
\usepackage{algorithmic}

\usepackage{natbib}

\usepackage[table]{xcolor}
\definecolor{mydarkred}{rgb}{0.6,0,0}
\definecolor{mydarkgreen}{rgb}{0,0.6,0}
\usepackage[colorlinks,
linkcolor=mydarkred,
citecolor=mydarkgreen]{hyperref}
\usepackage{url}

\usepackage{pifont}

\newtheorem{theorem}{Theorem}
\newtheorem{lemma}{Lemma}

\newcommand{\argmin}{\operatornamewithlimits{argmin}}
\newcommand{\argmax}{\operatornamewithlimits{argmax}}

\title{Multi-Class Classification from Noisy-Similarity-Labeled Data}

\author{
	Songhua Wu$^{1}\thanks{Equal contributions.}$,
	Xiaobo Xia$^{1,2}\footnotemark[1]$,
	Tongliang Liu$^{1}$,
	Bo Han$^{4,5}$,\\
	{Mingming Gong$^3$, Nannan Wang$^2$, Haifeng Liu$^6$, Gang Niu$^4$}\\[1ex]
	$^1$University of Sydney;
	$^2$Xidian University;
	$^3$The University of Melbourne;\\
	$^4$RIKEN;
	$^5$Hong Kong Baptist University;
	$^6$Brain-Inspired Technology Co., Ltd.
}

\date{}

\begin{document}
\bibliographystyle{plainnat}

\maketitle

\begin{abstract}
	A similarity label indicates whether two instances belong to the same class while a class label shows the class of the instance. Without class labels, a multi-class classifier could be learned from similarity-labeled pairwise data by meta classification learning \citep{hsu2019multi}. However, since the similarity label is less informative than the class label, it is more likely to be noisy. Deep neural networks can easily remember noisy data, leading to overfitting in classification. In this paper, we propose a method for learning from only noisy-similarity-labeled data. Specifically, to model the noise, we employ a noise transition matrix to bridge the class-posterior probability between clean and noisy data. We further estimate the transition matrix from only noisy data and build a novel learning system to learn a classifier which can assign noise-free class labels for instances. Moreover, we theoretically justify how our proposed method generalizes for learning classifiers. Experimental results demonstrate the superiority of the proposed method over the state-of-the-art method on benchmark-simulated and real-world noisy-label datasets.
\end{abstract}

\newpage

\section{Introduction}
Supervised classification crucially relies on the amount of data and the accuracy of corresponding labels. Since the data volume grows very quickly while supervision information cannot catch up with its growth, weakly supervised learning (WSL) is becoming more and more prominent \citep{zhou2017brief, han2019deep, wang2019symmetric, li2017learning, li2018cross, krause2016unreasonable, khetan2017learning, hu2019weakly}. Among WSL, similarity-based learning is one of the hottest emerging problems \citep{bao2018classification,hsu2019multi}. Compared with class labels, similarity labels are usually easier to obtain \citep{bao2018classification}, especially when we encounter some sensitive issues, e.g., religion and politics. Take an illustrative example from Bao \textit{et al.} \citep{bao2018classification}: for sensitive matters, people often hesitate to directly answer ``What is your opinion on issue A?"; while they are more likely to answer ``With whom do you share the same opinion on issue A?". Intuitively, similarity information can not only alleviate embarrassment but also protect personal privacy to some degree.

Existing methods for similarity-based learning can be divided into two categories generally: semi-supervised clustering \citep{wagstaff2001constrained, xing2003distance} and weakly-supervised classification \citep{bao2018classification, shimada2019classification}. The first category utilizes pairwise similarity and dissimilarity data for clustering. For example, pairwise links were used as constraints on clustering \citep{li2009constrained}; Similar and dissimilar data pairs were used for metric learning, which learns a distance function over instances and can easily convert to clustering tasks \citep{niu2014information}. The second category aims at classification, which not only separates different clusters but also identifies which class each cluster belongs to. For example, similarity and unlabeled (SU) learning proposed an unbiased estimator for binary classification \citep{bao2018classification}; Meta classification learning (MCL) showed a method to learn a multi-class classifier from only similarity data \citep{hsu2019multi}.

All existing methods are based on the strong assumption that similarity labels are entirely accurate. However, similarity labels are hard to be fully accurate for many applications. For example, for some sensitive matters, people may not be willing to provide their real thoughts even when facing easy questions. It is commonly known that deep networks can memorize all the training data even there is noisy supervision, which tends to lead to the overfitting problem \citep{zhang2016understanding,zhong2019graph, li2019learning, yi2019probabilistic, zhang2019metacleaner, tanno2019learning, zhang2018deep}. Thus, if we directly employ the existing deep learning algorithms to deal with noisy similarity-based supervision, the test performance will inevitably degenerate because of overfitting. To the best of our knowledge, no pioneer work has been done to tackle the problem of binary classification with noisy similarity information, not to mention how to learn multi-class classifiers with theoretical guarantees.

\begin{figure}[!tb]
	\centering
	\includegraphics[width=0.5\linewidth]{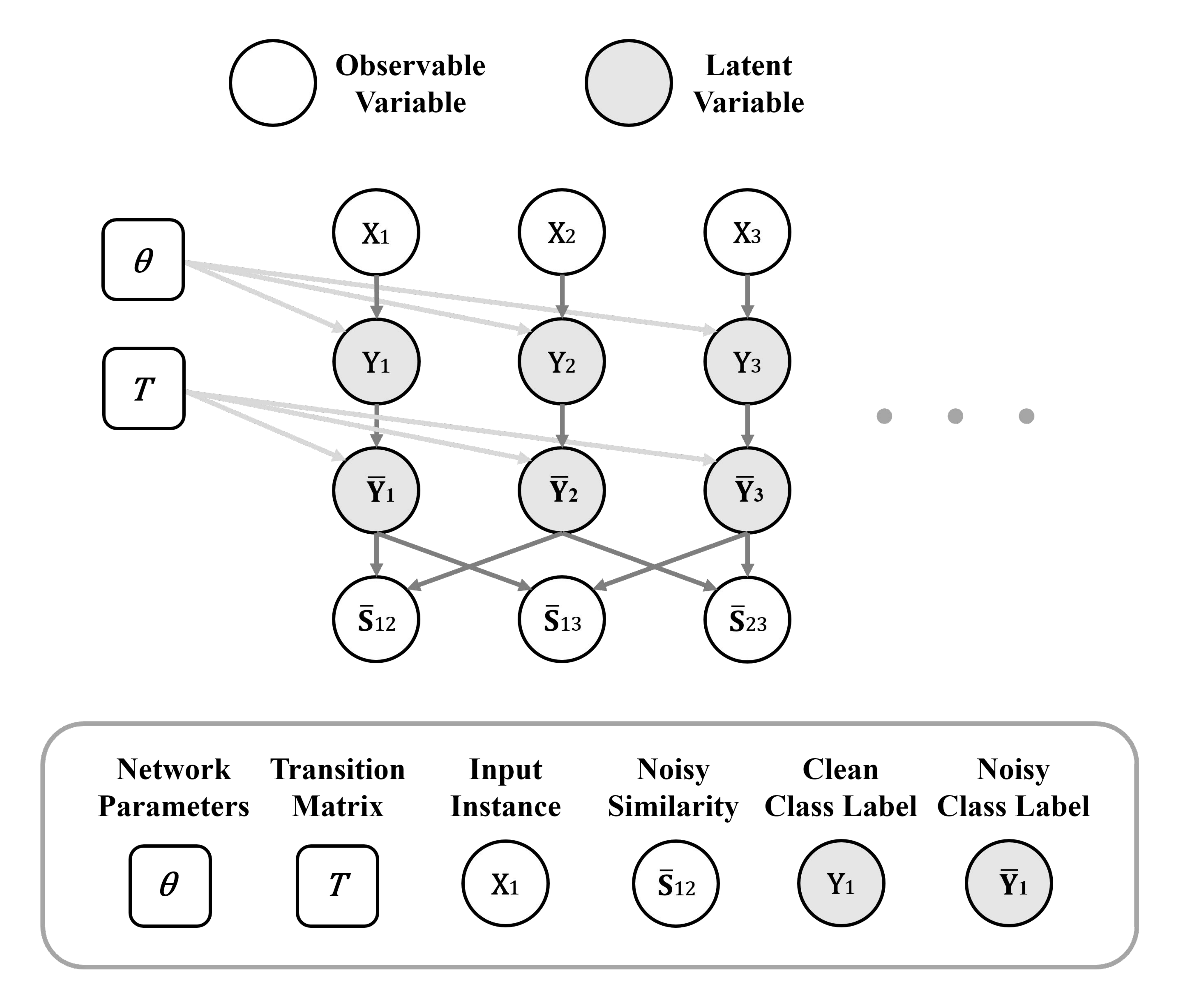}
	\caption{The assumed graphical representation for the proposed Multi-class classification with Noisy-Similarity-labeled data (called MNS classification), where $X_i$ denotes the input instance; $Y_i$ denotes the clean class label; $\bar{Y}_i$ denotes the noisy class label; $\bar{S}_{ii'}$ is noisy pairwise similarity supervision between $(X_i, \bar{Y}_i)$ and $(X_{i'}, \bar{Y}_{i'})$; $\theta$ denotes the neural network parameters; $T$ denotes the noise transition matrix. The latant variables are denoted by white circles and the observable variables are presented by grey circles.} \label{fig:setting}
\end{figure}

In this paper, we study the problem of how to learn a Multi-class classifier from Noisy-Similarity-labeled data, which is called MNS classification. Specifically, we assume that latent clean class labels $Y$ flip into latent noisy labels $\bar{Y}$, leading to noisy similarity labels $\bar{S}$. The corresponding graphical model, representing the interactions among variables, is shown in Figure \ref{fig:setting}. Based on this, we could further model the noise in the problem by using a transition matrix, 
i.e., $T_{ij}$ represents the probabilities that the clean class label $i$ flips into the noisy class label $j$ and $T_{ij}(X)=P(\bar{Y}=j|Y=i,X)$. We will show that under a mild assumption that \textit{anchor points} ( defined in \ref{sec:est_t}) exist in the training data, we can estimate the transition matrix by only employing noisy-similarity-labeled data. Then, we build a deep learning system for multi-class classification from only noisy-similarity-labeled data. Note that if a good classifier can be learned, the corresponding method can be easily extended to learn metrics or clusters, because accurate labels 
and similar and dissimilar pairs can be assigned by the good classifier. In other words, the proposed method can not only learn a classifier from noisy-similarity-labeled data but metrics and clusters. The contributions of this paper are summarized as follows:
\begin{itemize}
	\item We propose a deep learning system for multi-class classification to address the problem of how to learn from noisy-similarity-labeled data.
	
	\item We propose to model the noise by using the transition matrix based on a graphical model. We show that the transition matrix can be estimated from only noisy-similarity-labeled data. The effectiveness will be verified on both synthetic and real data.
	
	\item We theoretically establish a generalization error bound for the proposed MNS classification method, showing that the learned classifier will generalize well on unseen data.
	
	\item We empirically demonstrate that the proposed method can effectively reduce the side effect of noisy-similarity-labeled data. It significantly surpasses the baselines on many datasets with both synthetic noise and real-world noise \footnote{Datasets with real-world noise refer the noisy-similarity-labeled data where noisy similarity labels are generated using real-world data with label noise.}.
\end{itemize}

The rest of this paper is organized as follows. In Section 2, we formalize the MNS classification problem, and in Section 3, we propose the MNS learning and practical implementation. Generalization error bound is analysed in Section 4. Experimental results are discussed in Section 5. We conclude our paper in Section 6.

\section{Framing the \textbf{MNS classification} Problem}
\label{sec:2}

\textbf{Problem setup.}\ \
Let $\mathcal{D}$ be the distribution of a pair of random variables $(X,Y)\in\mathcal{X}\times[C]$, where $\mathcal{X}\subset \mathbb{R}^d$ and $d$ represents the dimension; $\mathcal{Y}=[C]$ is the label space and $[C] = \{1,\cdots,C\}$ is the number of classes. Our goal is to predict a label for any given instance $X\in\mathcal{X}$. Different from the traditional multi-class classification, in our setting, the class labels are not observable. Instead, we have noisy similarity labels $\bar{S}\in\{0,1\}$. The clean similarity labels $S$ indicate the similarities between examples, i.e., ${S}_{ii'} = \mathbbm{1}[Y_i = Y_{i'}]$ where $Y_i$ and $Y_{i'}$ denote the class labels for instances $X_i$ and $X_{i'}$. For noisy similarity labels, some of them are identical to the clean similarity labels, but some are different and we do not know which of them are clean. To the best of our knowledge, no existing work has discussed how to learn with the noisy similarity labels. We would like to review how the state-of-the-art work learns a classifier from the clean similarity labels.

\textbf{MCL classification} \citep{hsu2019multi}.\ \
Meta classification learning (MCL) utilizes the following likelihood to explain the similrity-based data
\begin{align} \label{eq:ccl_likelihood}
\mathcal{L}(\theta ; X, Y, S) &= P(X, Y, S; \theta) = P(S|Y)P(Y|X; \theta)P(X). 
\end{align}
By introducing an independence assumption: $S_{ii'} \perp S \setminus \{ S_{ii'}\} | X_{i},X_{i'}$ \citep[Appendix D]{hsu2019multi}, in other words, $S_{ii'}$ and $S \setminus \{ S_{ii'}\}$ are independent to each other given $X_{i}$ and $X_{i'}$; they can simplify the likelihood expression as
\begin{align}
\mathcal{L} (\theta ; X, S)&\approx \prod_{i,i'} \Big( \sum_{Y_i=Y_{i'}} \mathbbm{1}[S_{ii'}=1] P(Y_i|X_i;\theta) P(Y_{i'}|X_{i'};\theta)  \nonumber \\
& \quad + \sum_{Y_i \neq Y_{i'}} \mathbbm{1}[S_{ii'}=0] P(Y_i|X_i;\theta) P(Y_{i'}|X_{i'};\theta) \Big). \label{eq:ccl_likelihood_b}
\end{align}

Then taking a negative logarithm on Equation \ref{eq:ccl_likelihood_b}, the final loss function can be derived as
\begin{align}
L_{meta} (\theta)  &= - \sum_{i,i'} S_{ii'} \log (g(X_i;\theta)^T g(X_{i'};\theta)) \nonumber \\
& \quad + (1-S_{ii'})\log (1-g(X_i;\theta)^T g(X_{i'};\theta)), \label{eq:mcl_modify}
\end{align}
where $g(X_i;\theta) = P(Y_i | X_i; \theta)$, which can be learned from a neural network.

However, class label noise is ubiquitous in our daily life \citep{kaneko2019label, hu2019noise, zhong2019unequal, acuna2019devil, lee2018cleannet, tanaka2018joint, wang2018iterative}, not to mention the weaker supervision: similarity labels. The performance of classifiers will get worse if we still use the state-of-the-art methods designed for clean similarity labels. This motivates us to find a novel algorithm for learning from noisy-similarity-labeled data.

\section{MNS Learning}
\label{sec:others}
In this section, we propose a method for multi-class classification from noisy-similarity-labeled data.

\subsection{Modeling noise in the supervision}
To learn from the noisy-similarity-labeled data, we should model the noise. To model the noise, we introduce a graphic model in Figure \ref{fig:setting} to describe the interactions among variables, where only input instances $X$ and noisy similarity labels $\bar{S}$ are observed while both clean class labels $Y$ and noisy class labels $\bar{Y}$ are latent. Rather than modeling the similarity-label noise directly, we assume that noise first occurs on latent class labels and as a consequence, similarity labels turn to noisy ones, i.e., noisy similarity labels $\bar{S}_{ii'}\in\{0,1\}$ indicate the similarities between noisy examples, and $\bar{S}_{ii'} = \mathbbm{1}[\bar{Y}_i = \bar{Y}_{i'}]$. The assumption is reasonable. For example, in the sensitive matters, to hide one's thought on the question ``With whom do you share the same opinion on issue A?", people would like to randomly choose a fake opinion about the issue and answer the question conditioned on the fake opinion.

Specifically, to precisely describe label noise, we utilize a \textit{noise transition matrix} $T\in[0,1]^{C\times C}$ \citep{cheng2017learning}. The transition matrix is generally dependent on instances, i.e., $T_{ij}(X)=P(\bar{Y}=j|Y=i,X)$. Given only noisy examples, the instance-dependent transition matrix is non-identifiable without any additional assumption \citep{xia2019anchor}.
In this paper, we assume that given $Y$, $\bar{Y}$ is independent on instance $X$ and  $P(\bar{Y}=j|Y=i,X)=P(\bar{Y}=j|Y=i)$. This assumption considers the situations where noise relies only on the classes, which has been widely adopted in the class-label-noise learning community \citep{han2018co,xia2019anchor}. Empirical results on real-datasets verify the efficiency of the assumptions.

We denote by $\mathcal{D}_\rho$ the distribution of the noisy-similarity-labeled data $(X_i,X_{i'},\bar{S}_{ii'})$, and the classifier is supposed to be learned from a training sample drawn from $\mathcal{D}_\rho$.

\begin{figure*}[!t]
	\centering
	\includegraphics[width=0.98\textwidth]{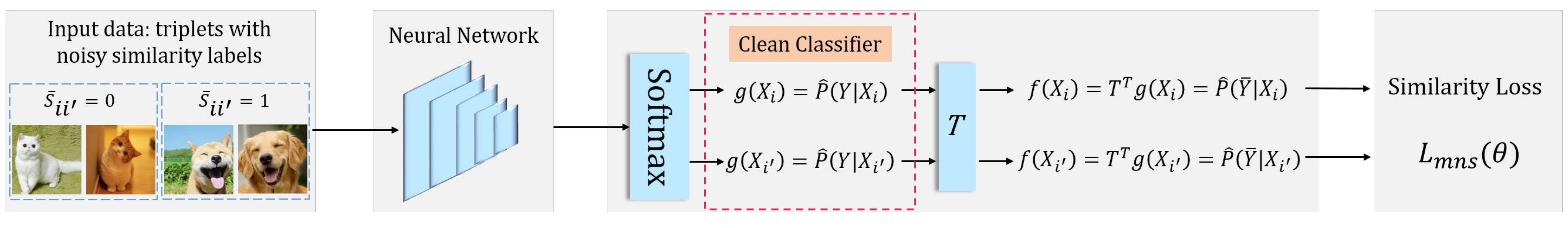}			
	\caption{An overview of the proposed method. We add a noise transition matrix layer to model the noise. By minimizing the proposed loss $L_{mns}(\theta)$, a classifier $g$ can be learned for assigning clean labels. The detailed structures of the Neural Network are provided in Section 5. Note that for the noisy similarity labels, some of them are correct and some are not. The similarity label for dogs is correct and the similarity label for cats is incorrect.
	}
	\label{fig:net}
\end{figure*}

\subsection{Likelihood-based estimator}\label{sec:3.2}

Intuitively, according to figure \ref{fig:setting}, we can explain the noisy-similarity-based data by using the following likelihood model
\begin{align}\label{eq:mns_likelihood}
\mathcal{L}(\theta ; X, Y, \bar{Y}, \bar{S})&= P(X, Y, \bar{Y}, \bar{S}; \theta) \nonumber
\\
&= P(\bar{S}|\bar{Y})P(\bar{Y}|Y)P(Y|X; \theta)P(X). 
\end{align}
In order to calculate the above likelihood, we have to marginalize the clean class label ${Y}$ and noisy class label $\bar{Y}$. Thanks to our proposed deep learning system (summarized in Figure \ref{fig:net}), $P(\bar{Y}|Y)$, modeled by a noise transition matrix $T$, could be learned only from noisy data (shown in Section \ref{sec:est_t}). Therefore, we only need to marginalize noisy class label $\bar{Y}$. With the independence assumption $S_{ii'} \perp S \setminus \{ S_{ii'}\} | X_{i},X_{i'}$, we can calculate the likelihood with the following expression
\begin{align}
\mathcal{L}(\theta ; X, Y, \bar{Y}, \bar{S})&\propto \sum_{\bar{Y}}\sum_{Y}P(\bar{S}|\bar{Y})P(\bar{Y}|Y)P(Y|X; \theta) \nonumber
\\
&= \prod_{i,i'} \Big( \sum_{\bar{Y}_i=\bar{Y}_{i'}} \mathbbm{1}[\bar{S}_{ii'}=1] \sum_{Y}P(\bar{Y_i}|Y)P(Y|X_i; \theta) \nonumber
\\
& \qquad \sum_{Y}P(\bar{Y_{i'}}|Y)P(Y|X_{i'}; \theta)  \nonumber \\
& \qquad + \sum_{\bar{Y}_i \neq \bar{Y}_{i'}} \mathbbm{1}[\bar{S}_{i{i'}}=0] \sum_{Y}P(\bar{Y_i}|Y)P(Y|X_i; \theta) \nonumber
\\
& \qquad \sum_{Y}P(\bar{Y_{i'}}|Y)P(Y|X_{i'}; \theta) \Big)  \nonumber \\
& = \prod_{i,i'} \Big( \sum_{\bar{Y}_i=\bar{Y}_{i'}} \mathbbm{1}[\bar{S}_{ii'}=1] P(\bar{Y}_i|X_i; \theta) P(\bar{Y}_{i'}|X_{i'}; \theta)  \nonumber \\
& \qquad + \sum_{\bar{Y}_i \neq \bar{Y}_{i'}} \mathbbm{1}[\bar{S}_{ii'}=0] P(\bar{Y}_i|X_i; \theta) P(\bar{Y}_{i'}|X_{i'}; \theta) \Big) \label{eq:final_likelihood} 
\end{align}
where the proportion relationship holds because $P(X)$ is constant for given $X$ such that can be omitted. Note that 
\begin{align}
P(\bar{Y}_i|X_i; \theta) &= \sum_{Y}P(\bar{Y_i}|Y)P(Y|X_i; \theta)\nonumber \\
& = \sum_{k=1}^C T_{ki}P(Y=k|X_i;\theta).
\end{align}
Let $g(X)=P({Y}|X;\theta)$ and $f(X)=P(\bar{Y}|X;\theta)$, we have
\begin{align}
f(X) = P(\bar{Y}|X;\theta) = T^\top P(Y|X;\theta) = T^\top g(X). \label{eq:g2f}
\end{align}
Then by taking a negative logarithm on Equation \ref{eq:final_likelihood} and substituting $P(\bar{Y}|X;\theta)$ with $f(X)$, we obtain the objective function of the proposed method, i.e.,
\begin{align}
L_{mns}(\theta ; X_i, X_{i'}, \bar{S}_{ii'}) &= - \sum_{i,{i'}} \log \Big(\sum_{\bar{Y}_i=\bar{Y}_{i'}} \mathbbm{1}[\bar{S}_{ii'}=1] P(\bar{Y}_i|X_i; \theta) P(\bar{Y}_{i'}|X_{i'}; \theta) \nonumber \\
&\quad + \sum_{\bar{Y}_i \neq \bar{Y}_{i'}} \mathbbm{1}[\bar{S}_{ii'}=0] P(\bar{Y}_i|X_i; \theta) P(\bar{Y}_{i'}|X_{i'}; \theta) \Big) \nonumber \\
&= - \sum_{i,{i'}} \bar{S}_{ii'} \log (f(X_i;\theta)^T f(X_{i'};\theta)) + \nonumber \\
&\qquad  (1-\bar{S}_{ii'})\log (1-f(X_i;\theta)^T f(X_{i'};\theta)). \label{eq:mns_criterion}
\end{align}

\begin{figure*}[!t]
	\begin{center}
		\subfigure[\textit{Similar example}]
		{\includegraphics[width=0.45\textwidth]{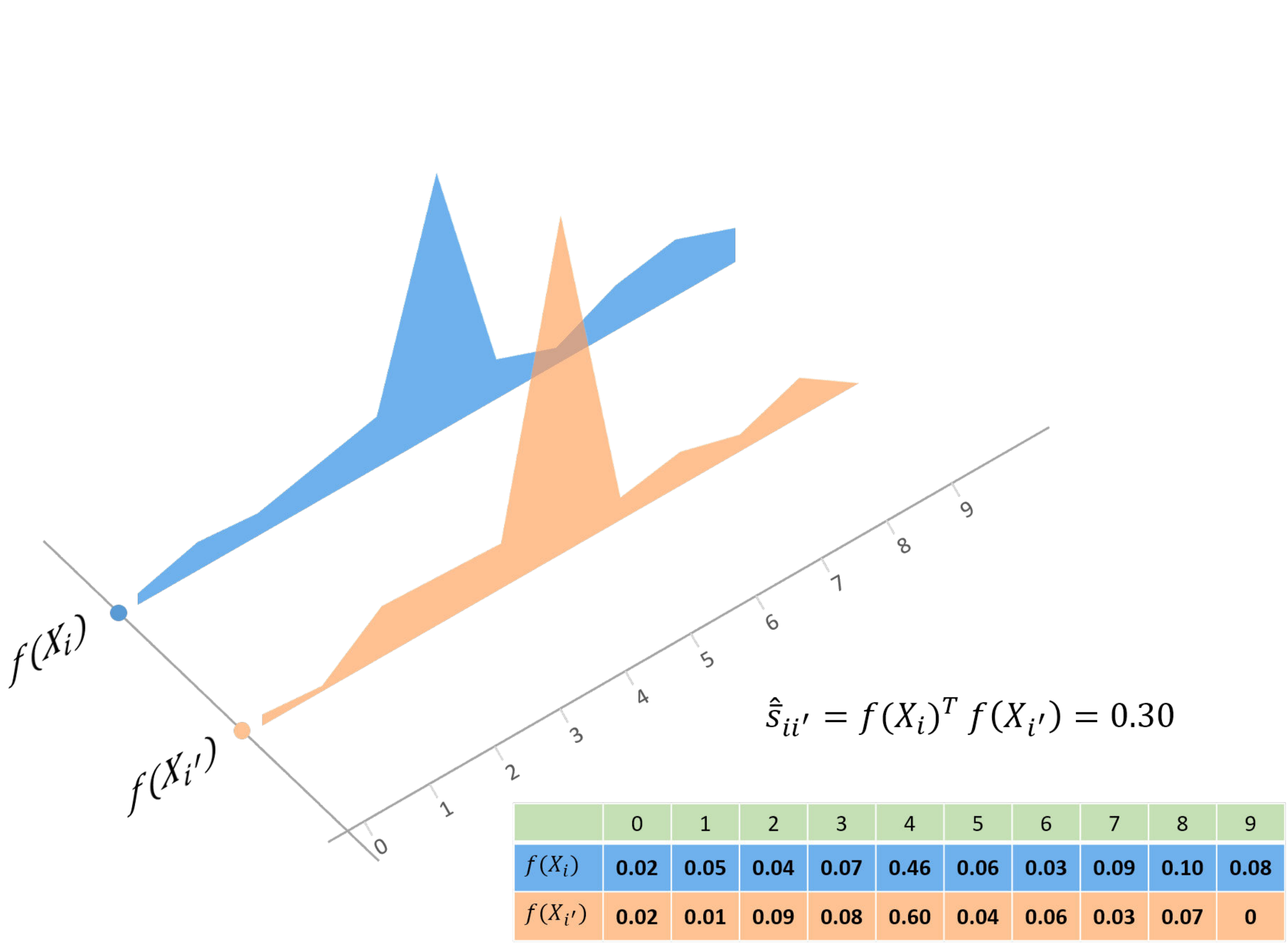}}
		\qquad
		\subfigure[\textit{Dissimilar example}]
		{\includegraphics[width=0.45\textwidth]{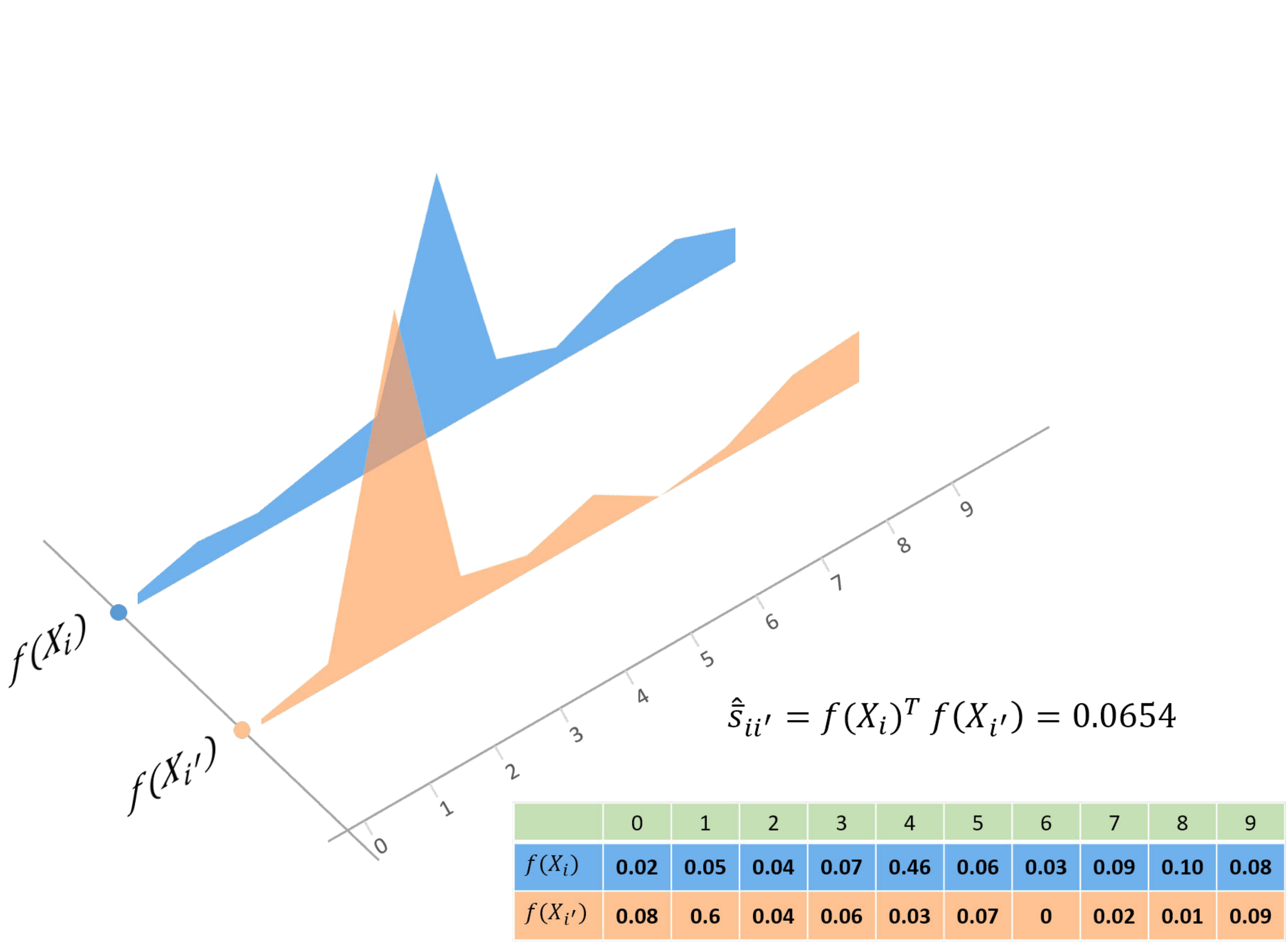}}
	\end{center}
	\caption{Examples of predicted noisy similarity. Assume class number is $10$; $f(X_i)$ and $f(X_{i'})$ are categorical distribution of instances $X_i$ and $X_{i'}$ respectively, which are shown above in the form of area charts. $\hat{\bar{S}}_{ii'}$ is the predicted similarity value between two instances, calculated by the inner product between two categorical distributions.}
	\label{fig:pre_simi}
\end{figure*}

Let us look inside Equation \ref{eq:mns_criterion}. Intuitively, $f(X;\theta)$ outputs the predicted noisy categorical distribution of instance $X$ and $f(X_i;\theta)^T f(X_{i'};\theta)$ is exactly the predicted noisy similarity, indicating the probability of data pairs belonging to the same noisy class. For clarity, we visualize the predicted noisy similarity in Figure \ref{fig:pre_simi}. If $X_i$ and $X_{i'}$ are predicted belonging to the same class, i.e., $\argmax_{m \in C} f_m(X_i;\theta) = \argmax_{n \in C} f_n(X_{i'};\theta)$, the predicted noisy similarity should be relatively high ($\hat{\bar{S}}_{ii'} = 0.30$ in Figure 3(a)). By contrast, if $X_i$ and $X_{i'}$ are predicted belonging to different classes, the predicted noisy similarity should be relatively low ($\hat{\bar{S}}_{ii'} = 0.0654$ in Figure 3(b)).

Further, let $\hat{\bar{S}}_{ii'} = f(X_i;\theta)^T f(X_{i'};\theta)$, denoting the predicted noisy similarity. Substituting $\hat{S}_{ii'}$ into Equation \ref{eq:mns_criterion}, $L_{mns}$ can convert into a binary cross-entropy loss version, i.e.,

\begin{align}
L_{mns}(\theta) = - \sum_{i,{i'}} \bar{S}_{ii'} \log \hat{\bar{S}}_{ii'} +   (1-\bar{S}_{ii'})\log (1-\hat{\bar{S}}_{ii'}). \label{eq:mns_ce_form}
\end{align}

Let us look inside Equation \ref{eq:mns_ce_form}. We could treat $\ell(\hat{\bar{S}}_{ii'},{\bar{S}}_{ii'})=- \bar{S}_{ii'} \log \hat{\bar{S}}_{ii'} +   (1-\bar{S}_{ii'})\log (1-\hat{\bar{S}}_{ii'})$ as the loss function denoting the loss of using $\hat{\bar{S}}_{ii'} $ to predict $\bar{S}_{ii'}$.
Then, our problem can be formulated in the traditional risk minimization framework \citep{mohri2018foundations}. The expected and empirical risks of employing estimator $f$ can be defined as
\begin{align}
{R}(f) = \mathbb{E}_{(X_i, X_{i'}, \bar{S}_{ii'})\sim {\mathcal{D}_\rho}}[{\ell}(f(X_i), f(X_{i'}), \bar{S}_{ii'})],
\end{align}
and
\begin{align}
{R}_n(f) = \frac{1}{n}\sum_{i=1}^n {\ell}(f(X_i), f(X_{i'}), \bar{S}_{ii'}),
\end{align}
where $n$ is training sample size of the noisy-similarity-labeled data.

The whole pipeline is summarized in Figure \ref{fig:net}. The softmax function outputs an estimator for the clean class posterior, i.e., $g(X)=\hat{P}(Y|X)$, where $\hat{P}(Y|X)$ denotes the estimated posterior. After the softmax layer, a noise transition matrix layer is added. According to Equation \ref{eq:g2f}, by pre-multiplying the transpose of the transition matrix, we can obtain a predictor $f(X)=\hat{P}(\bar{Y}|X)$ for the noisy class posterior, which can be further used to compute the prediction of the noisy similarity label, i.e., $\hat{\bar{S}}_{ii'}$. Therefore, by minimizing $L_{mns}$, as the training data goes to infinity, $\hat{\bar{S}}$ will converge to noisy similarity ${\bar{S}}$ and $f(X;\theta)$ will converge to the optimal classifier for predicting noisy class labels. Meanwhile, given the true transition matrix, $g(X)$ will converge to the optimal classifier for predicting clean class labels.

\subsection{Estimate noise transition matrix $T$} \label{sec:est_t}

However, the transition matrix is unknown. We will discuss how to estimate the transition matrix for the noisy-similarity-labeled data in this subsection.

\textit{Anchor points} \citep{liu2015classification,patrini2017making,yu2018learning} have been widely used to estimate the transition matrix for noisy-class-labeled data \citep{niu2018learning}. We illustrate that they can also be used to estimate the transition matrix for the noisy-similarity-labeled data. Specifically, an anchor point $x$ for class $y$ is defined as $P(Y=y|X=x)=1$ and $P(Y=y'|X=x)=0$, $\forall y'\in \mathcal{Y}\setminus \{y\}$. Let $x$ be an anchor point for class $i$ such that $P(Y=i|X=x)=1$ and for $k\neq i, P(Y=k|X=x)=0$. Then we have
\begin{align}\label{eq:anchor-estim}
P(\bar{Y}=j|X=x)&=\sum_{k=1}^C T_{kj}P(Y=k|X=x) \nonumber \\
&=T_{ij}P(Y=i|X=x)=T_{ij}.
\end{align}

Equation \ref{eq:anchor-estim} shows that given anchor points for each class and the noisy class posterior distribution, the transition matrix can be estimated. Note that the noisy class posterior can be estimated by $f(x)=\hat{P}(\bar{Y}|X)$ using the pipeline in Figure \ref{fig:net} without the transition matrix layer.
However, it is a bit strong to have access to anchor points. Instead, we assume that anchor points exist in the training data but unknown to us. Empirically, we select examples with the highest $\hat{P}(\bar{Y}=i|X=x)$ as anchor points for the $i$-th class.

\subsection{Implementation}
Given the true transition matrix, we can directly build a neural network as shown in Figure \ref{fig:net} to learn a multi-class classifier only from the noisy-similarity-labeled data. When the true transition matrix is unknown, we estimate it with the method proposed in Section \ref{sec:est_t} and then we can train the whole network as normal. The proposed algorithm is summarized in Algorithm \ref{alg:mns}.

\begin{algorithm}[t]
	{\bfseries Input}: noisy-similarity-labeled training data; noisy-similarity-labeled validation data.
	
	\textbf{Stage 1: Learn $\hat{T}$}
	
	1: Learn $f(X)=\hat{P}(\bar{Y}|X)$ by training the notwork in Figure \ref{fig:net} without the noise transition matrix layer;

	2: Estimate $\hat{T}$ according to Equation (\ref{eq:anchor-estim}) by using instances with the highest $\hat{P}(\bar{Y}|X)$ as anchor points;

	\textbf{Stage 2: Learn the classifier $g(X)=\hat{P}({Y}|X)$}
	
	3: Fix the transition matrix layer in Figure \ref{fig:net} by using the estimated transition matrix;

	4: Minimize $L_{mns}$ to learn $g$ and stop when $\hat{P}(\bar{Y}|X)$ corresponds the minimum classification error on the noisy validation set;

	{\bfseries Output}: $g$.
	\caption{MNS Learning Algorithm.}
	\label{alg:mns}
\end{algorithm}

\section{Generalization error}
In this section, we will theoretically analyze the generalization ability of the proposed method. Although it looks complex, we will show that it will generalize well.

Assume that the neural network has $d$ layers with parameter matrices $W_1,\ldots,W_d$, and the activation functions $\sigma_1,\ldots,\sigma_{d-1}$ are Lipschitz continuous, satisfying $\sigma_{j}(0)=0$. We denote by $h: X\mapsto W_d\sigma_{d-1}(W_{d-1}\sigma_{d-2}(\ldots \sigma_1(W_1X)))\in\mathbb{R}^C$ the standard form of the neural network. Then the output of the softmax function is defined as $g_i(X)=\exp{(h_i(X))}/\sum_{j=1}^{C}\exp{(h_j(X))}, i=1,\ldots,C$, and $f(X) = T^\top g(X)$ is the output of  the noise transition matrix layer. Let $\hat{f}=\argmax_{i\in\{1,\ldots,C\}}\hat{f}_i$ be the classifier learned from the hypothesis space $\mathcal{F}$ determined by the  neural network, i.e., $\hat{f}=\argmin_{f\in \mathcal{F}}{R}_{n}(f)$. Note that the risks are defined in Section \ref{sec:3.2}.

\begin{theorem} \label{thm:bound}
	Assume the parameter matrices $W_1,\ldots,W_d$ have Frobenius norm at most $M_1,\ldots, M_d$, and the activation functions are 1-Lipschitz, positive-homogeneous, and applied element-wise (such as the ReLU). Assume the transition matrix is given, and the instances are upper bounded by $B$, i.e., $\|X\|\leq B$ for all X, and the loss function $\ell(\hat{\bar{S}}_{ii'},{\bar{S}}_{ii'})$ is upper bounded by $M$\footnote{The assumption holds because deep neural networks will always regulate the objective to be a finite value and thus the corresponding loss functions are of finite values.}. Then, for any $\delta>0$, with probability at least $1-\delta$,
	\begin{align}
	{R}(\hat{f})- {R}_n(\hat{f}) \leq \frac{2BC(\sqrt{2d\log2}+1)\Pi_{i=1}^{d}M_i}{\sqrt{n}}+M\sqrt{\frac{\log{1/\delta}}{2n}}.
	\end{align}
\end{theorem}
A detailed proof is provided in Appendix.

Theorem \ref{thm:bound} implies that if the training error is small and the training sample size is large, the expected risk ${R}(\hat{f})$ of the learned classifier for noisy classes will be small. If the transition matrix is well estimated, the learned classifier for the clean class will also have a small risk according to Equation \ref{eq:g2f}. This theoretically justifies why the proposed method works well. In the experiment section, we will show that the transition matrices will be well estimated and that the proposed method will significantly outperform the baselines.

\section{Experiments}
In this section, we empirically investigate the performance of noise transition matrix estimation and the proposed method for MNS classification on three synthetic noisy datasets and two real-world noisy datasets.

\subsection{Experiments on synthetic noisy datasets}
\textbf{Datasets.}\ \
We synthesize noisy-similarity-labeled data by employing three widely used datasets, i.e., \textit{MNIST} \citep{lecun1998mnist}, \textit{CIFAR-10}, and \textit{CIFAR-100} \citep{krizhevsky2009learning}. \textit{MNIST} has $28 \times 28$ grayscale images of 10 classes including 60,000 training images and 10,000 test images. \textit{CIFAR-10} and \textit{CIFAR-100} both have $32 \times 32 \times 3$ color images including 50,000 training images and 10,000 test images. \textit{CIFAR-10} has 10 classes while \textit{CIFAR-100} has 100 classes. For all the three benchmark datasets, we leave out 10\% of the training examples as a validation set, which is for model selection. 

\textbf{Noisy similarity labels generation.}\ \
First, we artificially corrupt the class labels of training and validation sets according to noise transition matrices. Specifically, for each instance with clean label $i$, we replace its label by $j$ with a probability of $T_{ij}$. After that, we assign data pairs $((X_i,\bar{Y_i}),(X_{i'},\bar{Y_{i'}}))$ noisy similarity labels $\bar{S}_{ii'}$ and remove $\bar{Y_i}$ and $\bar{Y_{i'}}$. In this paper, we consider the symmetric noisy setting defined in Appendix. Noise-0.5 generates severe noise which means almost half labels are corrupted while Noise-0.2 generates slight noise which means around 20\% labels are corrupted. 


\textbf{Baselines.}\ \
We compare our proposed method with state-of-the-art methods and conduct all the experiments with default parameters by PyTorch on NVIDIA Tesla V100. Specifically, we compare with the following two algorithms: 
\begin{itemize}
	
	\item Meta Classification Likelihood (MCL) \citep{hsu2019multi}, which is the state-of-the-art method for multi-classification from clean-similarity-labeled data. 
	\item KLD-based Contrastive Loss (KCL) \citep{Hsu16_KCL}, which is a strong baseline. It uses Kullback–Leibler divergence to mesure the distance between two distributions. 
\end{itemize}

\textbf{Network structure.}\ \
For \textit{MNIST}, we use LeNet. For \textit{CIFAR-10}, we use pre-trained ResNet-32. For \textit{CIFAR-100}, we use VGG8. For all networks, as shown in Figure \ref{fig:net}, the output number of the last fully connected layer is set to be the number of classes. We add a noise transition matrix layer after the softmax. Since the loss functions of MNS, MCL and KCL are designed for instance pairs, a pairwise enumeration layer \citep{Hsu18_L2C} is adapted before calculating the loss.

\textbf{Optimizer.}\ \
We follow the optimization method in \citep{patrini2017making} to learn the noise transition matrix $\hat{T}$. To learn $g$, we use the Adam optimizer with initial learning rate 0.001.
On \textit{MNIST}, the batch size is 128 and the learning rate decays every 10 epochs by a factor of 0.1 with 30 epochs in total. On \textit{CIFAR-10}, the batch size is also 128 and the learning rate decays every 40 epochs by a factor of 0.1 with 120 epochs in total. On \textit{CIFAR-100}, the batch size is 1000 and the learning rate drops at epoch 80 and 160 by a factor of 0.1 with 200 epochs in total.

\begin{table*}[!t]
	\begin{center}
		\caption{Average Means and Standard Deviations (Percentage) of Classification Accuracy over 5 trials on \textit{MNIST}. KCL, MCL and MNS only have access to noisy similarity labels. Specifically, MCL($\hat{T}$) denotes the method in which we estimate noise transition matrix first and then use the estimated $\hat{T}$ for training while MCL($T$) skips the first step and directly use the true noise transition matrix.} 
		\renewcommand{\arraystretch}{1.4} 
		\vspace{8pt}
		\label{tab:syn1}
		\scalebox{0.9}
		{
			\begin{tabular}{c|ccccc}
				\hline
				Noise & 0.2 & 0.3 & 0.4 & 0.5 & 0.6 \\ \hline
				\hline
				KCL & \textbf{99.20$\pm$0.02} & \textbf{99.06$\pm$0.05} & 95.97$\pm$3.65 & 90.61$\pm$0.78 & 85.20$\pm$4.69 \\ 
				MCL & 98.51$\pm$0.10 & 98.28$\pm$0.06 & 97.92$\pm$0.24 & 97.54$\pm$0.09 & 96.94$\pm$0.20 \\ 
				MNS($\hat{T}$) & 98.56$\pm$0.07 & 98.29$\pm$0.16 & \textbf{98.01$\pm$0.15} & \textbf{97.61$\pm$0.41} & \textbf{97.26$\pm$0.23} \\ \hline
				MNS(${T}$) & 98.75$\pm$0.07 & 98.69$\pm$0.11 & 98.32$\pm$0.09 & 98.18$\pm$0.13 & 94.48$\pm$4.49 \\ \hline
			\end{tabular}
		}
	\end{center}
\end{table*}

\begin{table*}[t]
	\begin{center}
		\caption{Average Means and Standard Deviations (Percentage) of Classification Accuracy over 5 trials on \textit{CIFAR10}.}
		\renewcommand{\arraystretch}{1.4} 
		\vspace{8pt}
		\label{tab:syn2}
		\scalebox{0.9}
		{
			\begin{tabular}{c|ccccc}
				\hline
				Noise & 0.2 & 0.3 & 0.4 & 0.5 & 0.6 \\ \hline
				\hline
				KCL & 19.14$\pm$1.27 & 17.67$\pm$2.15 & 18.58$\pm$1.28 & 17.96$\pm$3.41 & 15.14$\pm$1.67 \\ 
				MCL & 75.58$\pm$3.64 & 68.90$\pm$0.32 & 63.38$\pm$1.32 & 61.67$\pm$0.98 & 44.55$\pm$2.96 \\ 
				MNS($\hat{T}$) & \textbf{78.83$\pm$1.81} & \textbf{76.80$\pm$1.33} & \textbf{70.35$\pm$1.21} & \textbf{68.87$\pm$0.97} & \textbf{50.99$\pm$2.88} \\ \hline
				MNS(${T}$) & 82.42$\pm$0.37 & 77.42$\pm$0.46 & 70.71$\pm$0.33 & 69.28$\pm$0.41 & 40.24$\pm$0.61 \\ \hline 
			\end{tabular}
		}
	\end{center}
\end{table*}

\begin{table*}[!t]
	\begin{center}
		\caption{Average Means and Standard Deviations (Percentage) of Classification Accuracy over 5 trials on \textit{CIFAR100}.}
		\renewcommand{\arraystretch}{1.4} 
		\vspace{8pt}
		\label{tab:syn3}
		\scalebox{0.9}
		{
			\begin{tabular}{c|ccccc}
				\hline
				Noise & 0.2 & 0.3 & 0.4 & 0.5 & 0.6 \\ \hline
				\hline
				KCL & 13.32$\pm$1.57 & 7.982$\pm$0.57 & 5.406$\pm$0.15 & 3.738$\pm$0.45 & 3.208$\pm$0.55 \\ 
				MCL & 48.38$\pm$0.38 & 40.48$\pm$0.79 & 32.75$\pm$0.77 & 26.48$\pm$0.36 & 21.94$\pm$0.19 \\ 
				MNS($\hat{T}$) & \textbf{48.78$\pm$0.74} & \textbf{43.90$\pm$0.39} & \textbf{40.26$\pm$0.93} & \textbf{35.14$\pm$0.69} & \textbf{31.40$\pm$0.26} \\ \hline
				MNS(${T}$) & 51.95$\pm$0.44 & 48.97$\pm$0.25 & 46.45$\pm$1.00 & 42.01$\pm$0.78 & 36.50$\pm$0.45 \\ \hline 
			\end{tabular}
		}
	\end{center}
\end{table*}

\textbf{Results.}\ \
The results in Tables \ref{tab:syn1}, \ref{tab:syn2}, and \ref{tab:syn3} demonstrate the test accuracy and stability of four algorithms on three benchmark datasets. Overall, we can see that when similarity labels are corrupted, MNS($\hat{T}$) achieves the best performance among three similarity-based learning methods, approaching or even exceeding MNS(${T}$) which is given the true noise transition matrix. Specifically, On \textit{MNIST} and \textit{CIFAR10}, when the noise rates are high, MNS($\hat{T}$) performs better than MNS(${T}$). This should because that $\hat{T}$ and the networks are learned jointly as shown in Algorithm \ref{alg:mns}.

On \textit{MNIST}, when the noise rate is relatively low (under 0.4), KCL has the highest accuracy; MCL and MNS also perform well. Intuitively, compared with inner product, Kullback-Leibler divergence measures the similarity between two distributions better, but it may introduce bad local minima or small gradients for learning \citep{hsu2019multi} such that it has poor performances on more complex datasets or higher noise rate. For example, when the noise rate increases (beyond 0.3), the accuracy of KCL drops dramatically, falling form 99.06 at Noise-0.3 to 85.20 at Noise-0.6. By contrast, MNS and MCL are more robust to noise. Both methods decrease slightly as the noise rate rises while our method is always a little better than the state-of-the-art method MCL.

On \textit{CIFAR-10} and \textit{CIFAR-100}, there is a significant decrease in the accuracy of all methods and our method achieves the best results across all noise rate, i.e., at Noise-0.6, MNS gives an accuracy uplift of about 6.5\% and 10\% on \textit{CIFAR-10} and \textit{CIFAR-100} respectively compared with the state-of-the-art method MCL.

\subsection{Experiments on real-world noisy datasets}
\textbf{Datasets.}\ \
We verify the effectiveness of the proposed method on two real-word datasets with noisy supervision, i.e., \textit{Clothing1M} \citep{xiao2015learning} and \textit{Food-101} \citep{bossard14}. Specifically,
\textit{Clothing1M} has 1M images with real-world noisy labels and additional 50k, 14k, 10k images with clean labels for training, validation and testing. We only use noisy training set in training phase and leave out 10\% as validation set for model selection and test our model on 10k testing set. \textit{Food-101} consists of 101 food categories, with 101,000 images. For each class, 250 manually reviewed clean test images are provided as well as 750 training images with real-world noise. For \textit{Food-101}, we also leave out 10\% for validation. In particular, we use \textit{Random Crop} and \textit{Random Horizontal Flip} for data augmentation. Since datasets contain some amount of class label noise already, we do not need to corrupt the labels artificially. We generate noisy-similarity-labeled data by using the noisy-class-labeled data directly.

\textbf{Baselines.}\ \ The same as the synthetic experiment part.

\textbf{Network structure and optimizer.}\ \
For all experiments, we use pre-trained ResNet-50. On \textit{Clothing1M}, the batch size is 256 and the learning rate drops every 5 epochs by a factor of 0.1 with 10 epochs in total. On \textit{Food-101}, the batch size is 1000 and the learning rate drops at epoch 80 and 160 by a factor of 0.1 with 200 epochs in total. Other settings are the same as the synthetic experiment part. 

\begin{table}[h]
	\begin{center}
		\caption{Classification Accuracy on real-world noisy dataset \textit{Clothing1M}. }
		\vspace{8pt}
		\label{tab:real1}
		\renewcommand{\arraystretch}{1.4} 
		\scalebox{0.9}
		{
			\begin{tabular}{ccc}
				\hline
				KCL & MCL & MNS($\hat{T}$)  \\ \hline
				9.49 & 66.20 & \textbf{67.50}\\ \hline
			\end{tabular}
		}
	\end{center}	
	\vspace{-18pt}
\end{table}

\begin{table}[!b]
	\begin{center}
		\caption{Classification Accuracy on real-world noisy dataset \textit{Food-101}.}
		\vspace{8pt}
		\label{tab:real2}
		\renewcommand{\arraystretch}{1.4} 
		\scalebox{0.9}
		{
			\begin{tabular}{ccc}
				\hline
				KCL & MCL & MNS($\hat{T}$)  \\ \hline
				30.17 & 48.08 & \textbf{71.18}\\ \hline	
			\end{tabular}
		}
	\end{center}	
\end{table}

\textbf{Results.}\ \
From Table \ref{tab:real1} and \ref{tab:real2}, We can see that on \textit{Clothing1M}, MNS($\hat{T}$) achieves the best accuracy. On \textit{Food-101}, MNS($\hat{T}$) also performs distinguishedly, uplifting about 23\% in accuracy compared with MCL. Specifically, the gap between MCL and MNS($\hat{T}$) is huge in Table \ref{tab:real2} while is not in Table \ref{tab:real1}. Let us review the definition of similarity-labeled data: if two instances belong to the same class, they will have similarity label $S = 1$, otherwise $S = 0$. That is to say, for a $k$-class dataset, only around $\frac{1}{k}$ of similarity-labeled data has similarity labels $S = 1$, and the rest $1 - \frac{1}{k}$ has similarity labels $S = 0$. For \textit{Clothing1M} (Table \ref{tab:real1}), the $k = 14$. For \textit{Food-101} (Table \ref{tab:real2}), the $k = 101$. Therefore, the generated similarity-labeled data from \textit{Food-101} is much more unbalanced than that from \textit{Clothing1M}. As a result, the baseline performed badly on \textit{Food-101}, making the gap huge in Table 5.

\subsection{Noise transition matrix estimation}
To estimate $T$, we first learn the noisy predictor $f(X)=\hat{P}(\bar{Y}|X)$. For each dataset, the network and optimizer remain the same as above but the noise transition matrix layer is exclude. $T$ is then estimated using the method proposed in Section \ref{sec:est_t}.

\begin{figure*}[h]
	\centering
	\subfigure[\textit{True transition matrix}]
	{\includegraphics[width=0.3\textwidth]{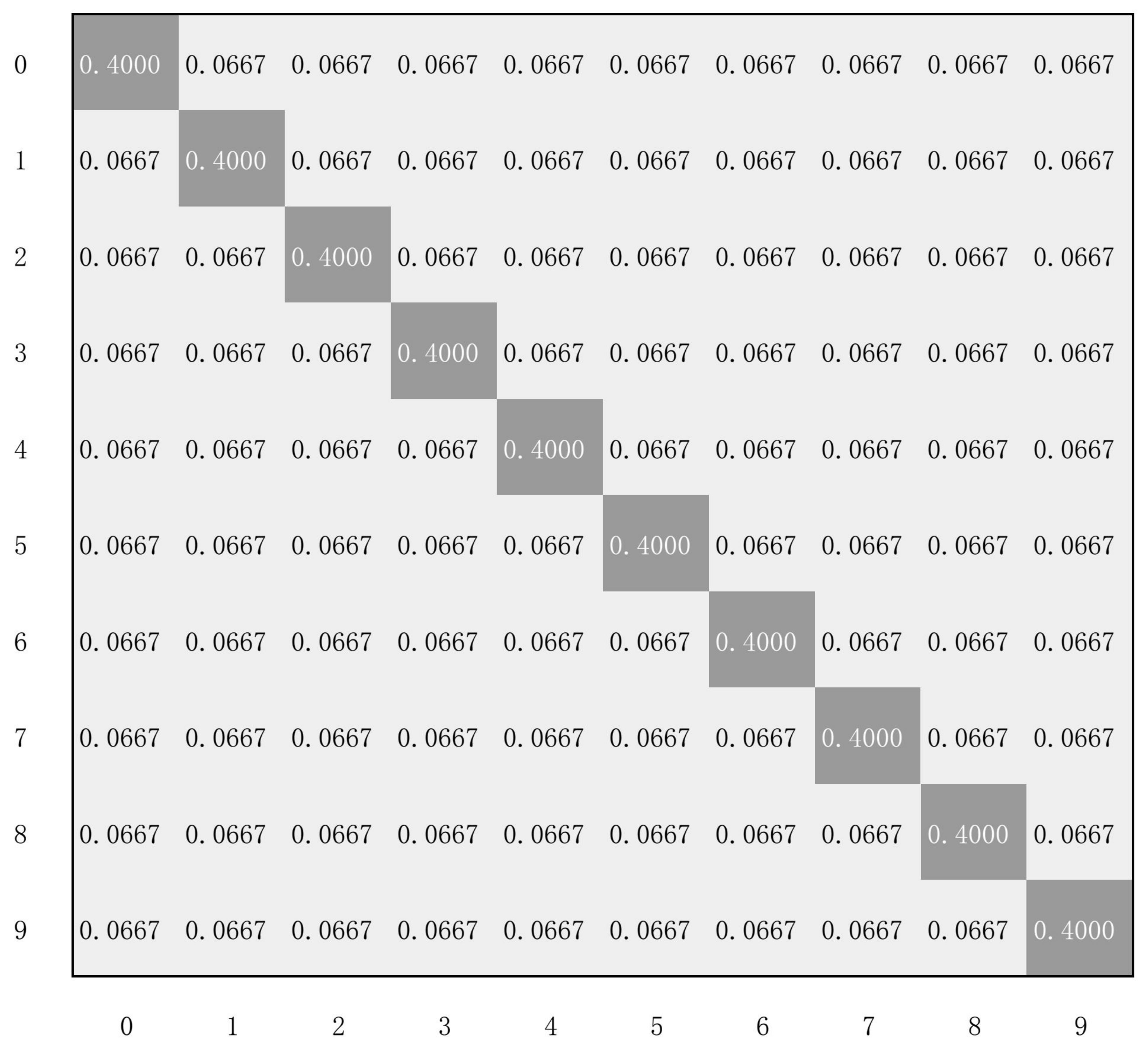}}
	\quad
	\subfigure[\textit{Estimated transition matrix on \textit{MNIST}}]
	{\includegraphics[width=0.3\textwidth]{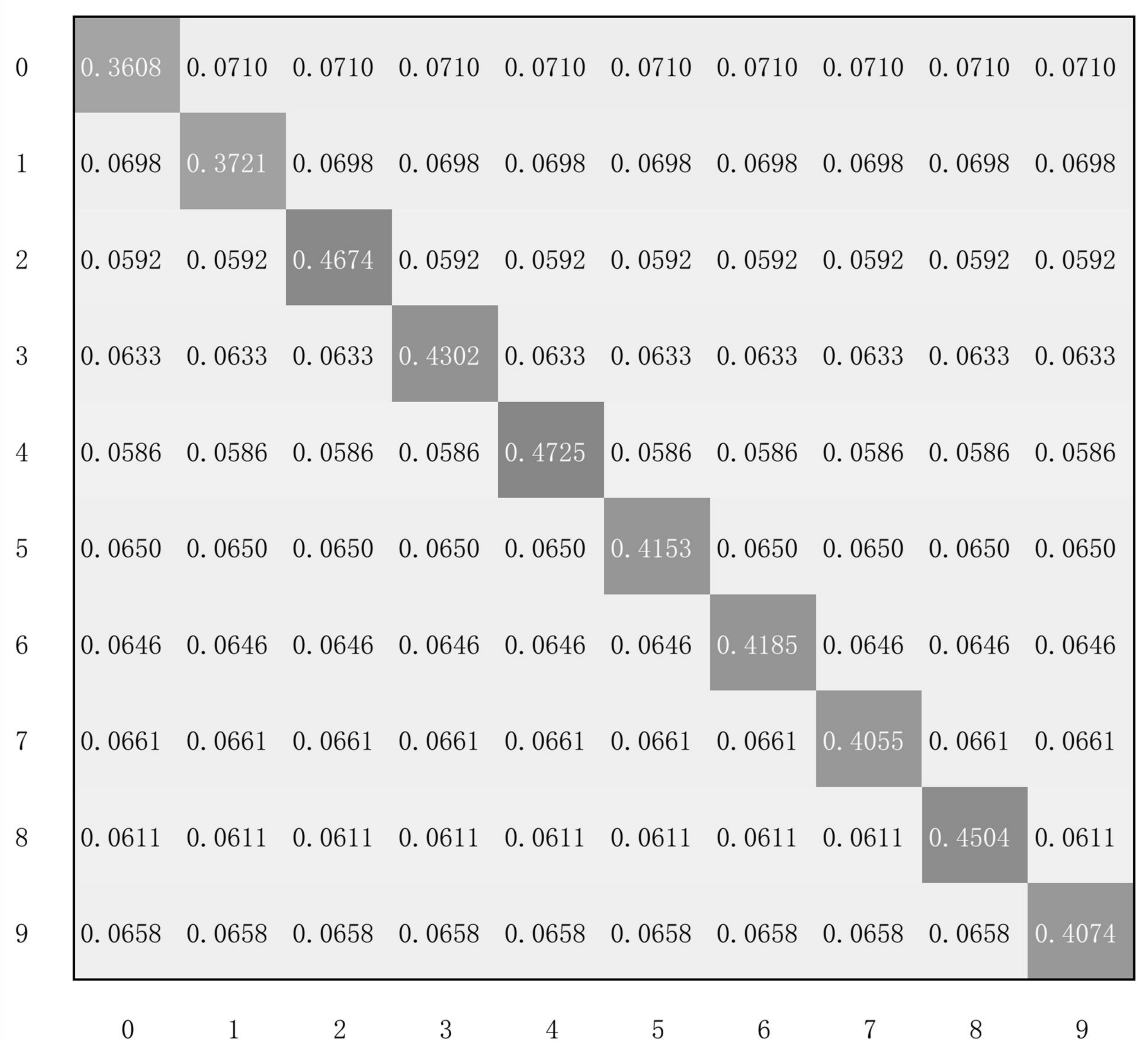}}
	\quad
	\subfigure[\textit{Estimated transition matrix on \textit{CIFAR-10}}]
	{\includegraphics[width=0.3\textwidth]{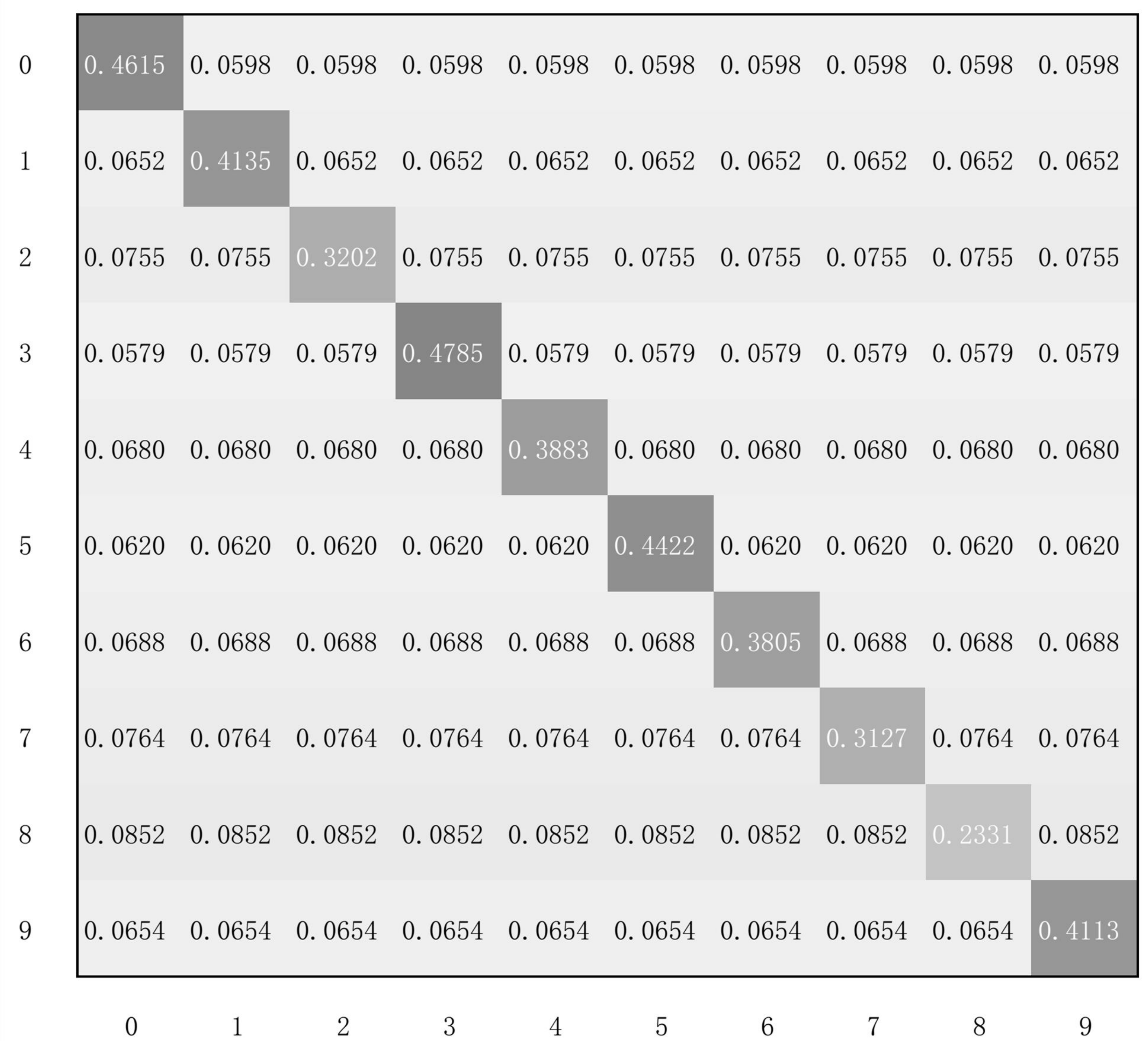}}
	\caption{True transition matrix $T$ at Noise-0.6 and corresponding $\hat{T}$ of two datasets with 10 classes: \textit{MNIST} and \textit{CIFAR-10}.}
	\label{fig:10_T}
\end{figure*}

\begin{figure*}[h]
	\begin{center}
		\subfigure[\textit{True noise transition matrix}]
		{\includegraphics[width=0.3\textwidth]{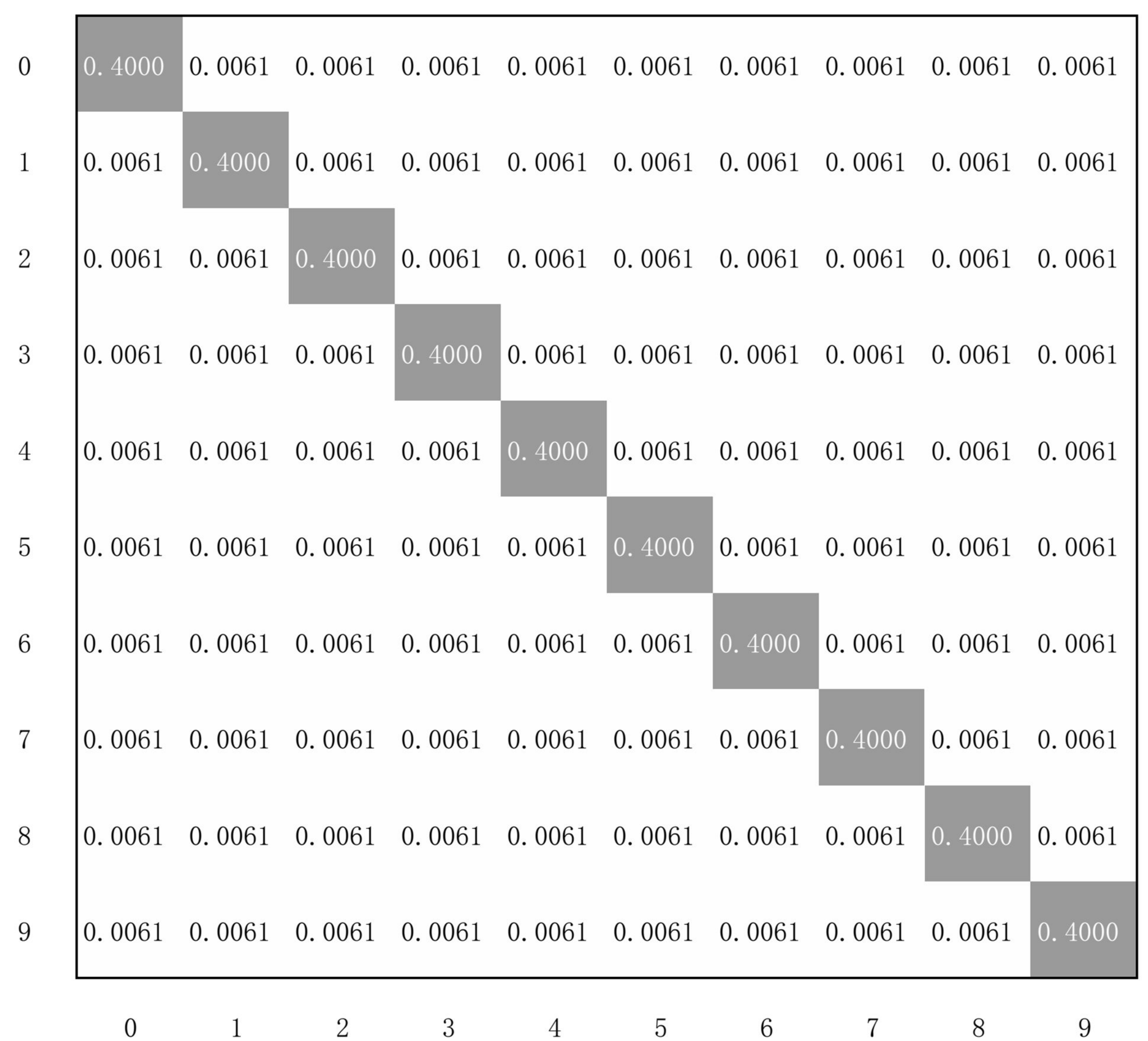}}
		\qquad
		\subfigure[\textit{Estimated transition matrix on \textit{CIFAR-100}}]
		{\includegraphics[width=0.3\textwidth]{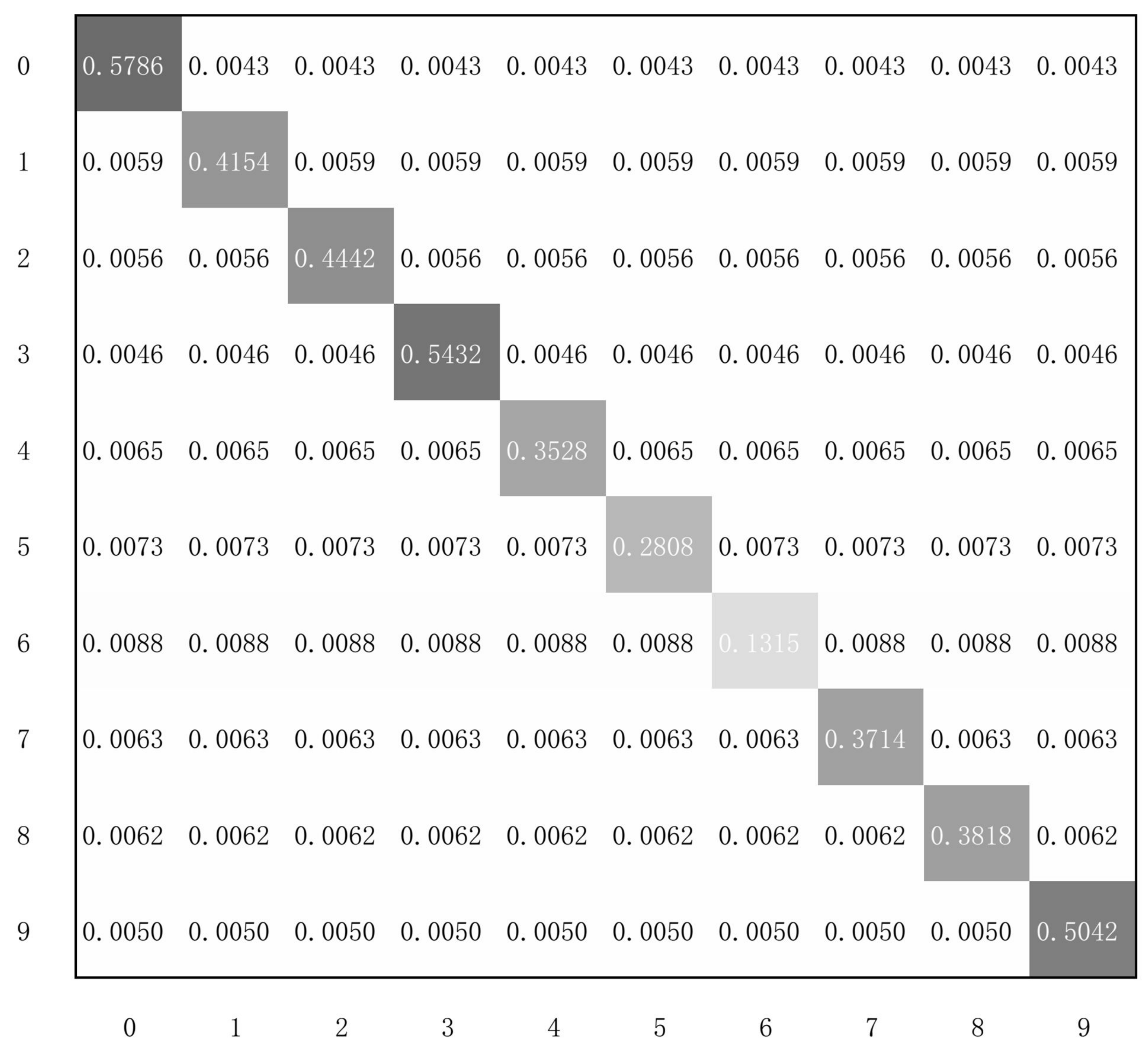}}
	\end{center}
	\caption{True transition matrix $T$ at Noise-0.6 and corresponding $\hat{T}$ of \textit{CIFAR-100}. Note that we only show the first 10 rows and columns of the matrix.}
	\label{fig:100_T}
\end{figure*}

Here we only show the estimated transition matrices of three synthetic noisy datasets because we have the exact values of the true transition matrices such that we could assess the estimation accuracies. Estimated transition matrices of real-world noisy datasets are provided in Appendix. From Figure \ref{fig:10_T} and \ref{fig:100_T}, we can see that transition matrices estimated with the proposed method are very close to the true one. By employing the calculation method of estimation error as $\epsilon = ||T-\hat{T}||_1/||T||_1$, \textit{MNIST}, \textit{CIFAR-10} and \textit{CIFAR-100} achieve 0.0668, 0.1144 and 0.1055 in error respectively.

\section{Conclusion}
This paper proposes a noisy-similarity-based multi-class classification algorithm (called MNS classification) by designing a novel deep learning system exploiting only noisy-similarity-labeled data. MNS classification provides an effective way for making predictions on sensitive matters where it is difficult to collect high-quality data such that similarities with noise could be all the information available. The core idea is to model the noise in the latent noisy class labels by using a noise transition matrix while only noisy similarity labels are observed. By adding a noise transition matrix layer in the deep neural network, it turns to robust to similarity label noise. We also present that noise transition matrix can be estimated in this setting. Experiments are conducted on benchmark-simulated and real-world label-noise datasets, demonstrating our method can excellently solve the above weakly supervised problem. In future work, investigating different types of noise for diverse real-life scenarios might prove important.

\clearpage
\bibliographystyle{plain}
\bibliography{bib}

\newpage
\section*{Appendices}

\appendix
\setcounter{theorem}{1}

\section{Proof of Theorem 1}
We have defined
\begin{align}
{R}(f) = \mathbb{E}_{(X_i, X_{i'}, \bar{S}_{ii'})\sim {\mathcal{D}_\rho}}[{\ell}(f(X_i), f(X_{i'}), \bar{S}_{ii'})],
\end{align}
and
\begin{align}
{R}_n(f) = \frac{1}{n}\sum_{i=1}^n {\ell}(f(X_i), f(X_{i'}), \bar{S}_{ii'}),
\end{align}
where $n$ is training sample size of the noisy-similarity-labeled data.

First we bound the generalization error with Rademacher complexity \citep{bartlett2002rademacher}.
\begin{theorem}[\citep{bartlett2002rademacher}]\label{thm:rademacher}
	Let the loss function be upper bounded by $M$. Then, for any $\delta>0$, with the probability $1-\delta$, we have
	\begin{equation}
	\begin{aligned}
	\sup_{f\in \mathcal{F}}|{R}(f)-{R}_n(f)|\leq 2\mathfrak{R}_n({\ell}\circ\mathcal{F})+M\sqrt{\frac{\log{1/\delta}}{2n}},
	\end{aligned}
	\end{equation}
	where $\mathfrak{R}_n({\ell}\circ\mathcal{F})$ is the Rademacher complexity defined by
	\begin{align}
	\mathfrak{R}_n({\ell}\circ\mathcal{F})=\mathbb{E}\left[\sup_{f\in \mathcal{F}}\frac{1}{n}\sum_{i=1}^n\sigma_i{\ell}(f(X_i), f(X_{i'}), \bar{S}_{ii'})\right],
	\end{align}
	and $\{\sigma_1,\cdots,\sigma_n\}$ are Rademacher variables uniformly distributed from $\{-1,1\}$.
\end{theorem}

Before further upper bound the Rademacher complexity $\mathfrak{R}_n({\ell}\circ\mathcal{F})$, we discuss the special loss function and its \textit{Lipschitz continuity} w.r.t $h_j(X_i), j = \{1,\ldots,C\}$.
\begin{lemma} \label{lemma:Lipschitz}
	Given transition matrix $T$, loss function ${\ell}(f(X_i), f(X_{i'}), \bar{S}_{ii'})$ is 1-Lipschitz with respect to $h_j(X_i), j = \{1,\ldots,C\}$,
	\begin{align}
	&\left|\frac{\partial {\ell}(f(X_i), f(X_{i'}), \bar{S}_{ii'})}{\partial h_j(X_i)}\right|<1.
	\end{align}
\end{lemma}

Detailed proof of Lemma \ref{lemma:Lipschitz} can be found in Section \ref{pf:lem1}.

Based on Lemma \ref{lemma:Lipschitz}, we can further upper bound the Rademacher complexity $\mathfrak{R}_n({\ell}\circ\mathcal{F})$ by the following lemma.

\begin{lemma} \label{lemma:f2h}
	Given transition matrix $T$ and assume loss function ${\ell}(f(X_i), f(X_{i'}), \bar{S}_{ii'})$ is 1-Lipschitz with respect to $h_j(X_i), j = \{1,\ldots,C\}$, we have
	\begin{align}
	\mathfrak{R}_n({\ell}\circ\mathcal{F})&=\mathbb{E}\left[\sup_{f\in \mathcal{F}}\frac{1}{n}\sum_{i=1}^n\sigma_i{\ell}(f(X_i), f(X_{i'}), \bar{S}_{ii'})\right] \nonumber \\
	&\leq C\mathbb{E}\left[ \sup_{h\in H }\frac{1}{n}\sum_{i=1}^{n}\sigma_ih(X_i)\right],
	\end{align}
	where $H$ is the function class induced by the deep neural network.
\end{lemma}

Detailed proof of Lemma \ref{lemma:f2h} can be found in Section \ref{pf:lem2}.

The right hand part of the above inequality, indicating the hypothesis complexity of deep neural networks, can be bound by the following theorem.

\begin{theorem}[\citep{golowich2017size}]\label{thm:network}
	Assume the Frobenius norm of the weight matrices $W_1,\ldots,W_d$ are at most $M_1,\ldots, M_d$. Let the activation functions be 1-Lipschitz, positive-homogeneous, and applied element-wise (such as the ReLU). Let $X$ is upper bounded by B, i.e., for any $X$, $\|X\|\leq B$. Then,
	\begin{align}
	\mathbb{E}\left[ \sup_{h\in H }\frac{1}{n}\sum_{i=1}^{n}\sigma_ih(X_i)\right]\leq \frac{B(\sqrt{2d\log2}+1)\Pi_{i=1}^{d}M_i}{\sqrt{n}}.
	\end{align}
\end{theorem}

Combining Lemma \ref{lemma:Lipschitz},\ref{lemma:f2h}, and Theorem \ref{thm:rademacher}, \ref{thm:network}, Theorem 1 is proven.

\subsection{Proof of Lemma 1}\label{pf:lem1}

Recall that
\begin{align}
{\ell}(f(X_i), f(X_{i'}), \bar{S}_{ii'} = 1) &= -\log (f(X_i)^T f(X_{i'})). \nonumber \\
&= -\log \big((T^Tg(X_i))^T(T^Tg(X_{i'}))\big), 
\end{align}
where
\begin{align}
g(X_i) &= [g_1(X_i),\ldots,g_c(X_i)] \nonumber \\ 
&= \left[\left(\frac{\exp(h_{1}(X))}{\sum_{j=1}^c \exp(h_j(X))}\right),\ldots,\left(\frac{\exp(h_{c}(X))}{\sum_{i=j}^c \exp(h_j(X))}\right)\right]^T.
\end{align}

Take the derivative of ${\ell}(f(X_i), f(X_{i'}), \bar{S}_{ii'} = 1)$ w.r.t. $h_j(X_i)$, we have
\begin{equation} \label{derivative1}
\begin{aligned}
\frac{\partial {\ell}(f(X_i), f(X_{i'}), \bar{S}_{ii'} = 1)}{\partial h_j(X_i)} = \frac{\partial {\ell}(f(X_i), f(X_{i'}), 1)}{\partial f(X_{i'})^T f(X_i)}  \frac{\partial f(X_{i'})^T f(X_i)}{\partial f(X_i)}   \frac{\partial f(X_i)}{\partial g(X_i)} \frac{\partial g(X_i)}{\partial h_j(X_i)},
\end{aligned}
\end{equation}
where
\begin{align*}
&\frac{\partial {\ell}(f(X_i), f(X_{i'}), \bar{S}_{ii'} = 1)}{\partial f(X_{i'})^T f(X_i)} = -\frac{1}{f(X_{i'})^T f(X_i)} \nonumber \\ 
&\frac{\partial f(X_{i'})^T f(X_i)}{\partial f(X_i)} = f(X_{i'})^T \nonumber \\
&\frac{\partial f(X_i)}{\partial g(X_i)} = T^T \nonumber \\
&\frac{\partial g(X_i)}{\partial h_j(X_i)} = g'(X_i) = [g'_1(X_i),\ldots,g'_c(X_i)].
\end{align*}
Note that the derivative of the softmax function has some properties, i.e.,
if $m \neq j$, $g'_m(X_i) = -g_m(X_i)g_j(X_i)$ and if $m = j$, $g'_j(X_i) = (1-g_j(X_i))g_j(X_i)$.

We denote by $Vector[m]$ the $m-th$ element in $Vector$ for those complex vectors. Because $\ 0<g_m(X_i)<1, \forall m\in\{1,\ldots,c\}$ and $T_{ij} > 0, \forall i,j\in\{1,\ldots,c\}$, we have
\begin{align}
&g'_m(X_i) \leq |g'_m(X_i)| < g_m(X_i), &\forall m\in\{1,\ldots,c\};  \\ 
&T^Tg'(X_i)[m] < T^T|g'(X_i)|[m] < T^Tg(X_i)[m], &\forall m\in\{1,\ldots,c\}.	
\end{align}
Since $0 < f_m(X_{i'})^T < 1, \forall m\in\{1,\ldots,c\}$, similarly we have
\begin{align}
&f(X_{i'})^TT^T|g'(X_i)| < f(X_{i'})^TT^Tg(X_i) = f(X_{i'})^T f(X_i).
\end{align}
Therefore, 
\begin{align}
&\left|\frac{\partial {\ell}(f(X_i), f(X_{i'}), \bar{S}_{ii'} = 1)}{\partial h_j(X_i)}\right| \nonumber \\
&= \left|\frac{\partial {\ell}(f(X_i), f(X_{i'}), 1)}{\partial f(X_{i'})^T f(X_i)}  \frac{\partial f(X_{i'})^T f(X_i)}{\partial f(X_i)}   \frac{\partial f(X_i)}{\partial g(X_i)} \frac{\partial g(X_i)}{\partial h_j(X_i)}\right| \nonumber \\
&= \left|-\frac{f(X_{i'})^T T^T g'(X_i)}{f(X_{i'})^T f(X_i)}\right| \nonumber \\
&\leq \left| \frac{f(X_{i'})^T T^T |g'(X_i)|}{f(X_{i'})^T f(X_i)}\right| \nonumber \\
&< \left| \frac{f(X_{i'})^T f(X_i)}{f(X_{i'})^T f(X_i)}\right| = 1.
\label{eq:L0}
\end{align}
Similarly, we can proof 
\begin{align}
&\left|\frac{\partial {\ell}(f(X_i), f(X_{i'}), \bar{S}_{ii'} = 0)}{\partial h_j(X_i)}\right|<1. \label{eq:L1}
\end{align}
Combining Eq.\ref{eq:L0} and Eq.\ref{eq:L1}, we obtain
\begin{align}
&\left|\frac{\partial {\ell}(f(X_i), f(X_{i'}), \bar{S}_{ii'})}{\partial h_j(X_i)}\right|<1. \label{eq:L}
\end{align}

\subsection{Proof of Lemma 2}\label{pf:lem2}

\begin{align*}
&\mathbb{E}\left[\sup_{f\in \mathcal{F}}\frac{1}{n}\sum_{i=1}^n\sigma_i{\ell}(f(X_i), f(X_{i'}), \bar{S}_{ii'})\right]  \\
&= \mathbb{E}\left[\sup_{g }\frac{1}{n}\sum_{i=1}^{n}\sigma_i{\ell}(f(X_i), f(X_{i'}), \bar{S}_{ii'})\right] \\
&= \mathbb{E}\left[\sup_{\argmax\{h_1,\ldots,h_C\}}\frac{1}{n}\sum_{i=1}^{n}\sigma_i{\ell}(f(X_i), f(X_{i'}), \bar{S}_{ii'})\right] \\
&= \mathbb{E}\left[\sup_{\max\{h_1,\ldots,h_C\} }\frac{1}{n}\sum_{i=1}^{n}\sigma_i{\ell}(f(X_i), f(X_{i'}), \bar{S}_{ii'})\right] \\
&\leq \mathbb{E}\left[ \sum_{k=1}^C \sup_{h_k\in H }\frac{1}{n}\sum_{i=1}^{n}\sigma_i{\ell}(f(X_i), f(X_{i'}), \bar{S}_{ii'})\right] \\
&= \sum_{k=1}^C \mathbb{E}\left[ \sup_{h_k\in H }\frac{1}{n}\sum_{i=1}^{n}\sigma_i{\ell}(f(X_i), f(X_{i'}), \bar{S}_{ii'})\right] \\
&\leq C\mathbb{E}\left[ \sup_{h_k\in H }\frac{1}{n}\sum_{i=1}^{n}\sigma_ih_k(X_i)\right] \\
&=C\mathbb{E}\left[ \sup_{h\in H }\frac{1}{n}\sum_{i=1}^{n}\sigma_ih(X_i)\right],
\end{align*}
where the first three equations hold because given $T$, $f, g$ and $\max\{h_1,\ldots,h_C\}$ give the same constraint on $h_j(X_i), j = \{1,\ldots,C\}$; the sixth inequality holds because of the Lemma \citep{ledoux2013probability}.

\section{Definition of transition matrix}
Symmetric noisy setting is defined as follows, where \textit{C} is the number of classes.
\begin{align}
\text{Noise-$\rho$:}
& \quad
T =
\begin{bmatrix}
1-\rho & \frac{\rho}{C-1} & \dots  & \frac{\rho}{C-1} & \frac{\rho}{C-1}\\
\frac{\rho}{C-1} & 1-\rho & \frac{\rho}{C-1} & \dots & \frac{\rho}{C-1}\\
\vdots &  & \ddots &  & \vdots\\
\frac{\rho}{C-1} & \dots & \frac{\rho}{C-1} & 1-\rho & \frac{\rho}{C-1}\\
\frac{\rho}{C-1} & \frac{\rho}{C-1} & \dots  & \frac{\rho}{C-1} & 1-\rho
\end{bmatrix}.
\end{align}

\section{Estimation of transition matrix on real-world noisy datasets}
Here we show the estimated transition matrices of \textit{Clothing1M} and the first ten classes of \textit{Food-101}. For \textit{Clothing1M}, we use additional 50k images with clean labels to learn the transition matrix such that the left $\hat{T}$ in Figure 1 is very close to the true one. The right $\hat{T}$ in Figure 1 was estimated only from noisy-similarity-labeled data, which learned most of the features of true transition matrix. For \textit{Food-101}, both $\hat{T}$ was estimated from noisy-labeled data. From Figure 2 we can see that the result close to the result which verifies the effectiveness of our method.
\begin{figure}[h]	\label{fig:T_clothing}
	\begin{center}
		\includegraphics[width=0.83\textwidth]{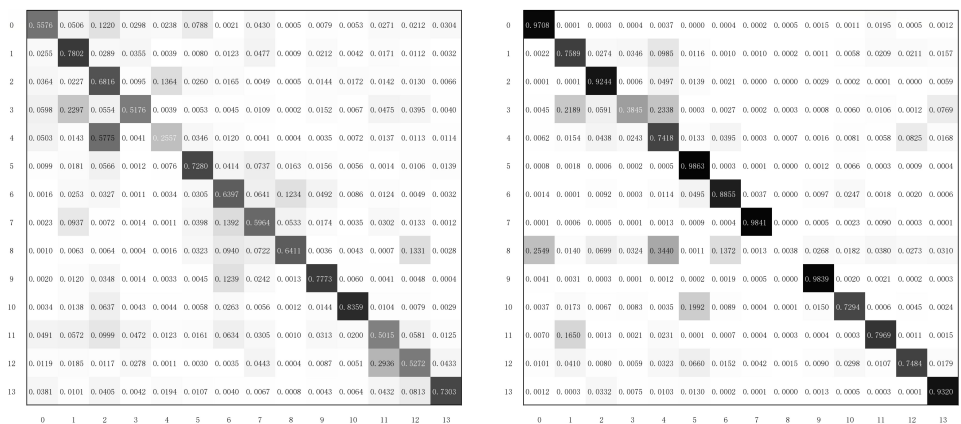}			
	\end{center}
	\caption{$\hat{T}$ of \textit{Clothing1M}; the one in the left hand is $\hat{T}$ estimated from class labels, the one in the right hand is $\hat{T}$ estimated from noisy similarity labels.}
\end{figure}

\begin{figure}[h] 	\label{fig:T_food}
	\begin{center}
		\includegraphics[width=0.83\textwidth]{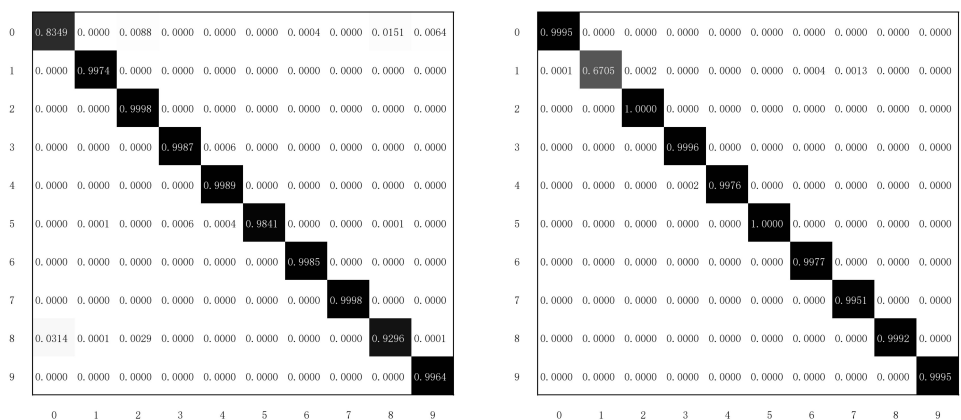}			
	\end{center}
	\caption{$\hat{T}$ of the first ten classes of \textit{Food-101}; the one in the left hand is $\hat{T}$ estimated from class labels, the one in the right hand is $\hat{T}$ estimated from noisy similarity labels}
\end{figure}

\end{document}